\documentclass[9pt,journal,twoside,web]{ieeecolor}
\usepackage[outdir=./]{epstopdf}
\usepackage{generic}
\usepackage{cite}
\usepackage{amsmath,amssymb,amsfonts}
\usepackage{algorithm}
\usepackage{algpseudocode}

\usepackage{graphicx}
\usepackage{textcomp}
\usepackage{algpseudocode}

\usepackage{booktabs}

\def\BibTeX{{\rm B\kern-.05em{\sc i\kern-.025em b}\kern-.08em
    T\kern-.1667em\lower.7ex\hbox{E}\kern-.125emX}}
\begin{document}
\title{Semi-Autonomous Laparoscopic Robot Docking with Learned Hand-Eye Information Fusion}

\author{
Huanyu Tian$^{1,2}$,
Martin Huber$^{1}$, Christopher E. Mower$^{1}$, Zhe Han$^{1,3}$, \\Changsheng Li$^{2}$,\IEEEmembership{Senior Member, IEEE}, Xingguang Duan$^{2,3}$, and Christos Bergeles$^{1}$, \IEEEmembership{Senior Member, IEEE}%
\thanks{$^{1}$H.~Tian, M.~Huber, C.~E.~Mower, Z.~Han, and C.~Bergeles are with the School of Biomedical Engineering \& Imaging Sciences, King’s College London, UK. Corresponding author:
        {\tt\small christos.bergeles@kcl.ac.uk}}%
\thanks{$^{2}$H.~Tian, C.~Li, and X.~Duan are with the School of Mechatronical Engineering, Beijing Institute of Technology, Beijing 100081, China, and with the Key Laboratory of Biomimetic Robots and Systems, Beijing Institute of Technology, Ministry of Education, China}
\thanks{$^{3}$  Z.~Han  and X.~Duan are with School of Medical Technology, Beijing Institute of Technology, Beijing 100081, China, and with the Key Laboratory of Biomimetic Robots and Systems, Beijing Institute of Technology, Ministry of Education, China}
\thanks{This work was partially supported by the Wellcome/EPSRC Centre for Medical Engineering [WT203148/Z/16/Z], EPSRC [EP/Y024281/1], the European Union's Horizon 2020 research and innovation program under grant agreement No 101016985 (FAROS project), and Beijing Institute of Technology under Chasing Dream Abroad Scholarship. For the purpose of Open Access, the Author has applied a CC BY public copyright license to any Author Accepted Manuscript version arising from this submission.
}
}

\maketitle

\begin{abstract}

In this study, we introduce a novel shared-control system for key-hole docking operations, combining a commercial camera with occlusion-robust pose estimation and a hand-eye information fusion technique. This system is used to enhance docking precision and force-compliance safety. To train a hand-eye information fusion network model, we generated a self-supervised dataset using this docking system. After training, our pose estimation method showed improved accuracy compared to traditional methods, including observation-only approaches, hand-eye calibration, and conventional state estimation filters.
In real-world phantom experiments, our approach demonstrated its effectiveness with reduced position dispersion (1.23$\pm$0.81 mm vs. 2.47 $\pm$ 1.22 mm) and force dispersion (0.78$\pm$0.57 N vs. 1.15$\pm$0.97 N) compared to the control group. These advancements in semi-autonomy co-manipulation scenarios enhance interaction and stability. The study presents an anti-interference, steady, and precision solution with potential applications extending beyond laparoscopic surgery to other minimally invasive procedures.
\end{abstract}


\section{Introduction}
\label{Section1}

A preliminary phase of laparoscopic surgery, known as \textit{docking}, precedes the main procedure and involves the insertion of an endoscope~\cite{ChrisCARS} or surgical instruments through trocars \cite{LongqianARDock}, 
i.e. percutaneous cannulas that allow access to the insufflated abdomen \cite{heemskerk2007robot}.
Safe docking is identified as a key consideration during operating room setup~\cite{malik2022robotic}. Therefore, our interest is in semi-autonomy in docking as a means to improve safety and reduce the burden on theatre staff.
Our objective is to create a holistic framework comprising a controller and a state estimator. This framework is designed to enable semi-autonomous trocar docking, serving as the initial stage in advancing collaborative human/robot laparoscopic surgery.

Semi-autonomy, an approach for control that combines input from a human and an autonomous agent,
has been proposed and applied in areas such as 
healthcare \cite{attanasio2021autonomy}, 
manufacturing and construction\cite{Mower2022An}, and 
humanoid control \cite{Marion17}.
Seen as a stepping stone towards full autonomy of surgical robots 
it can reduce the burden on a surgeon whilst improving safety (e.g. by modulating online motion constraints~\cite{Moccia20}) and 
address ethical considerations~\cite{yang2017medical} (i.e. by subverting the issue of accountability in the fully automated case).
Methods implementing semi-autonomy effectively merge the innate reasoning and manipulation abilities of humans with the high-precision capabilities of robots \cite{tian2023virtual}. In the system, humans can offer superior situational awareness to enable intervention for supervision and unexpected-event handling \cite{tianSharedControl}.


\subsection{Related Works}

Docking is a variation of the classical ``peg-in-hole'' problem:
the robot attempts to insert the tip of an object, fixed to the end-effector (the endoscope in our case), into a hole (i.e. the trocar for us).
Typically, the hole is positioned in a fixed location. However, in our case, we have the additional challenge that the hole (trocar) is free to pivot and is sensitive to the force applied by the instrument~\cite{ChrisCARS}.
Due to this sensitivity, the insertion direction may vary, requiring continuous realignment of the robot end-effector. 
As the trocar is directly linked to the patient, ensuring safety is imperative. 
This encourages the adoption of an optimization-based semi-autonomy strategy.
Optimization-based semi-autonomy~\cite{Mower2022An} enables us to model optimal system behavior by a scalar-valued cost function, whilst ensuring safety by imposing motion limitations as constraints.
Such an approach allows the system to be directed by humans, providing the system with contextual awareness at a human level. 
Simultaneously, we can impose strict limits on our system model, enabling us to guarantee safety.
Moreover, with the robot's kinematics/dynamic model in mind, we can formulate a control system within the manipulator task space to enhance usability for the user and alleviate the cognitive workload.

The initial stages of docking involve identifying the location of the entry target and determining the appropriate method to guide the robot for alignment and insertion of its end-effector into the target.
Marker-based optical tracking systems for pose estimation like NDI Polaris in \cite{ZheHanRespiratoryMotion} 
remain the most prevalent methods for high-accuracy pose estimation due to their ability to offer reliable and robust vision features.
Several alternative approaches have been explored, such as
retroreflective markers \cite{wu2019accurate},
RGBD cameras \cite{liebmann2024automatic},
bespoke black-and-white markers/tags \cite{wang2023cylindertag}, and 
even tags with corners and wireless IMU \cite{pflugi2018augmented}. 
Markerless pose estimation of trocars using a synthetic dataset \cite{dockingsystemDehghani}, and 
external sensing of passive arm stands \cite{yip2019vision} have been proposed. The marker-based method is the most common, efficient and stable method among these. 
However, unfortunately, marker-based methods or any line-of-sight method for pose estimation are affected by occlusions on any features of their patterns \cite{JiaoleWang}.

\begin{figure}[t]
	\centering
	\includegraphics[width=\columnwidth]{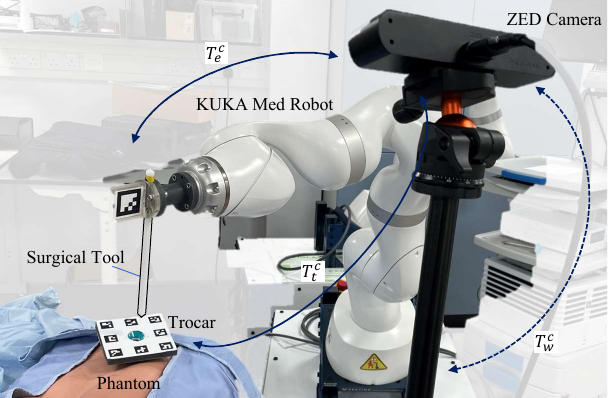}
	  \caption{
   The docking system leverages RGB cameras, the robot, and the trocar with markers to conduct the laparoscopic instrument insertion procedure. The pose from the camera to the target (trocar in our case) is $T_{t}^{c}$. To register the robot, the pose from the camera to the end-effector $T_{e}^{c}$ can be observed to infer the hand-eye calibration results i.e. $T_{c}^{w}$. However, in this context, the camera's movement, the occlusion of the two markers, and key-point detection failures could cause measurements' outliers.
   Note, the dashed line represents a transformation that is initially unknown without calibration.
   }
	  \label{figure1}
\end{figure}

Considering robot motion control, one approach is to follow a ``\textit{look and move}'' policy\cite{CHANG2018102}. 
This method relies on high quality pose estimation derived from a view of the trocar. 
An eye-in-hand autonomous trocar docking system has been proposed to achieve alignment and insertion of a tool in vitreoretinal surgery\cite{dockingsystemDehghani}, its focus being the estimation of the trocar pose. 
Repeatable docking was showcased, but without force feedback~\cite{lee2022peg, compliantDocking} or hand-eye fusion~\cite{JingXuTII}, which are proven to be helpful in peg-in-hole task success rate. 
Although a high-end camera with low distortion and high resolution can be used in surgical settings, 
we empirically identified that a stable positioning error of under $2\,$mm is still not easy to achieve in a meter-level workspace using ``\textit{look and move}'' policies.
Also, typical patient physiological movements (e.g. breathing) and positional disturbances during the operation bring challenges and uncertainty in the pose estimation.

Another approach is visual servoing, which includes position-based visual servoing (PBVS) \cite{PBVSControl} and image-based visual servoing (IBVS) \cite{IBVSControl}. Contrary to ``\textit{look and move}'', visual servoing closes the loop of robot motion using continuously updated image information. 

In this study, we investigate the PBVS method as it explicitly represents motion goals (i.e. positions and orientations) in the common Cartesian space, rather than the image space.
This enables us to combine these motion goals, derived from vision, with human-interactive forces\footnote{External torques measured using the robot joint sensors after direct physical interaction between the human and robot.} representing guidance shared control and active task space motion constraints.
However, PBVS relies on real-time pose estimation and is therefore sensitive to the noise introduced when keypoints that inform target object pose are detected in the images. Such noise especially affects orientation estimation \cite{yang2023object}, and can cause unwanted robot motions.

As shown in Figure \ref{figure1}, the camera is not attached to the robot end-effector.
Rather, it is situated next to the robot orientated at the surgical scene.
When the camera 
is not rigidly attached to the robot end-effector, 
the pose of the camera with respect to the robot base frame is unknown.
In order to estimate this transformation we utilize a common hand-eye registration approach~\cite{HECAli01} in our pipeline.
Calibration is a common step in image-based navigation in surgery\cite{birkfellner1998calibration,jiang2022overview}.
In general, this approach has the disadvantage that the camera can not be moved after calibration. 
However, in surgical settings, this assumption is reasonable and is employed in current surgical procedures.


The common way to handle registration is hand-eye calibration \cite{handeyecaliDan}. Conventional hand-eye calibration needs more than 5 well-designed measurements offline and is affected by measurement noise \cite{zhang2017computationally} brought by pose estimation. Maximum errors could even reach $5\,$mm - $8\,$mm. In a surgical setup, the necessary use of small markers could lead to even greater errors, thereby further affecting performance \cite{WANG2023105127}. In addition, the error is also affected by the measurement's scale, i.e. only considering the errors brought from the camera model (like distortions or imperfect camera matrix), when the robot approaches to the target, the measurement's error of the distance between the robot and the target will reduce. Therefore, real-time correction\cite{DynamicVS} (i.e. information fusion between the robot's positions and the camera's observations) should be introduced to adaptively adjust the homogeneous transformation from the end effector to the target 
That is $T_{t}^{e} = ({T_{e}^{c}})^{-1} T_{t}^{c}$, shown in Fig.\ \ref{figure1},
where $T_a^b\in\mathbb{R}^{4\times 4}$ is the homogeneous transformation matrix representing the pose of the child frame $a$ in the parent frame $b$. 
Although the initial measurement error may be large, the error will reduce while the robot approaches the trocar. However, potential occlusions and outliers that might arise from wrong detection/estimation should also be addressed.

Introduction of a Kalman filter or one of its variants\cite{PBVSKF} offer two advantages. 
First, a Kalman filter can use adaptive weighting to balance the importance of observations and model-based predictions (robotic kinematics/dynamics in our case) \cite{OpticalLocalization}. 
Second, some variants of Kalman filters could reject the outliers measured by the sensors' failed measurements with internal distribution-based outlier detection \cite{OutlierRejection}. 
Their smooth nature is advantageous to robotic applications because it leads to low-frequency motions for the hardware to execute. 
On the down side, these filters have a large set of parameters that require manual tuning.

The Kalman filter is a Bayesian-based method that requires a model of the measurements' confidence as well as the model's confidence\cite{KFsurvey}. Deriving the measurements' confidence lacks reliable design guidelines, while a Kalman filter addressing non-linear dynamics may be hard to fine-tune \cite{ZHAO2020109184}. Also, the basic Kalman filter and its variants have limitations in handling non-linear, non-smooth, and even non-injective representations for orientation in SO3 like quaternions \cite{sola2017quaternion}. To address this issue, one can leverage the error states \cite{ESKFMP, ESKF_IMU} in filters, which mean the filter no longer tracks the state of the target but tracks changes in the target state instead. This, however, increases the difficulty of fine-tuning the filter and dealing with outliers. 

Data-driven methods can fine-tune filter performance through offline and online learning of the model's and filter's parameters \cite{krishnan2015deep}. A successful attempt to create a learning-based Kalman filter is the Kalman-Net \cite{revach2022kalmannet}. However, to the best of our knowledge, the KalmanNet has, so far, only been applied in simulations, and the datasets including training sets and validation sets are given by manually designed functions with artificial noise. 
We did not find work that has been carried out for surgical scenarios and in the context of hand-eye information fusion \cite{thiemjarus2012eye}. In addition, learning-based methods require a high-quality training dataset with ground truth that is subject itself to noise and imperfection. 
The dataset requires an overall docking setup including pose estimation, controller, and kinematics which simulates how the robot starts from an initial pose and ultimately docks into the trocar. This is a typical Markov process \cite{kurniawati2022partially} along the docking trajectory.

Overall, the existing work on laparoscopic docking applications has several notable limitations: First, the approach lacks real-time tracking ability due to the ``\textit{look-and-move}" policy and omitted force perception. Second, current approaches face challenges in achieving a straightforward and easily fine-tuned method for online outlier exclusion. Finally, there is still a research gap on learned filter network structure and the corresponding dataset addressing the hand-eye fusion (or calibration in real time) in the context of surgical docking.

\subsection{Contributions}
To train the data-driven filters, we use hand-eye calibration with offline filter-free trajectories. At execution time during the procedure, we only need to be able to see the marker and robot state and so we can move camera. In operation, the trained data-driven filters can handle the PBVS control and doesn't depend on hand-eye calibration.
In this work, we make the following contributions:
\begin{itemize}
\item A learning-based hand-eye information fusion method to ensure accurate positioning for trocar aligning and rejection of measurement outliers arising from failed pose estimation, occlusions, and accidental camera motion.
\item A marker-based state estimation method that establishes accurate ground truth, and a new hand-eye calibration approach that builds a dataset in a self-supervised fashion.
\item A robot-assisted docking system with a semi-autonomous control method that incorporates target tracking and compliant behavior.
\end{itemize}

To aid with replication of our results, the software and relevant hardware files will be made available online. 

\section{System Overview}
\label{Section2}

Our system, shown in Fig.~\ref{figure1}, includes a robotic manipulator (the KUKA LBR Med 7 R800) and RGB camera (ZED camera).
The camera is not required to be fixed during operation as it aims to obtain the transformation between the end of the robot and the trocar.
The camera provides feedback for the semi-autonomous visual servo control.

\subsection{Co-manipulation for docking}

To decrease the difficulty of doctors' operations and improve the interpretability of the system\cite{alonso2018system},
task-allocation distinction is applied to divide the task space into two subspaces for human-robot interaction (HRI) control, and environment-robot interaction control, respectively.
The co-manipulation for docking is decomposed into two parts: 
\textbf{(1)} alignment of the trocar axis and the tool, and
\textbf{(2)} insertion of the tool into the trocar.

The aligning task was governed by the robot itself (details shown in Sec.~ \ref{Section5}A and Sec.~\ref{Section5}B), with lateral translation and rotation constrained to align with the trocar in real time. As there is symmetry along the direction of rotation about the tool's axis, the controller does not need to align the remaining orientation. This property enables virtual fixture constraints in only one rotational direction rather than full SO3 space, which introduces the redundancy of the robot and therefore, increases the workspace of the robot. 

The surgeon manually controls the tool insertion speed through their guiding force: increased guiding force leads to a more rapid robot response (details shown in Sec.~\ref{Section5}C).


\begin{figure}[t]
	\centering
	\includegraphics[width=\columnwidth]{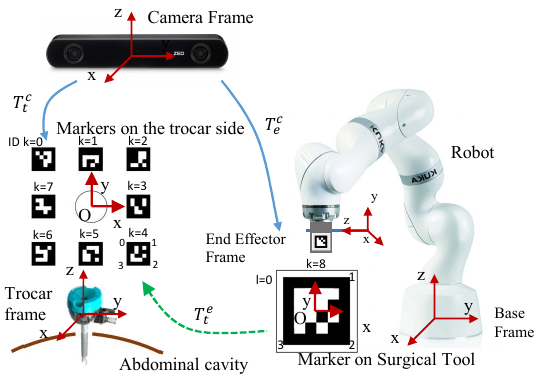}
	  \caption{Markers are attached on both the robot (end-effector) and the trocar. To reduce the chance of catastrophic occlusions during co-manipulation, a multi-marker ensemble is used. 
   The coordinate frame of ArUco markers planar board is considered the world coordinate frame.}
    \label{figure2}
\end{figure}

\subsection{Vision system and marker design}

We opt for marker-based pose estimation in the developed semi-autonomous docking system, as shown in Fig.\ \ref{figure2}. We use a single ArUco marker, with size of $45\,$mm, to track the pose of the robot end-effector in the camera frame. On the trocar side, marker design considered the inevitable occlusion from the tool during its insertion. To mitigate this, we created an array that includes $8$ discrete ArUco markers \cite{garrido2014automatic}, arranged clockwise on a planar board clamped around the trocar. The full set of markers and their corners is used in subsequent registration steps. The size of each marker is $20$\,mm.  All markers are generated using OpenCV (i.e. cv2.aruco.DICT\_4X4\_250).

\section{State Estimation for Docking}
\label{Section3}

Semi-autonomy, e.g. semi-autonomous docking in our case, entails estimating the values of the target's pose and Human-Robot Interaction (HRI) force/environment-robot interaction force, i.e. the system states.

\subsection{Pose estimation}
The OpenCV ArUco library can detect the ArUco marker corners in pixel coordinates, and order them by ArUco ID, $k$, and intrinsic corner point ID, $l$. Therefore, it is straightforward to create a set of correspondences $\hat{S}=\{\mathbf{u}_{k, l} \leftrightarrow \mathbf{p}_{k,l}\}$, where, for each corner ID, $l$, and marker ID, $k$, $\mathbf{u}_{k, l}$ are the corner coordinates in the $2$D input image and $\mathbf{p}_{k, l}$ are $3$D coordinates of the ArUco marker corner with respect to the coordinate frame of the planar board, which can be considered the world coordinate frame. Using $\hat{S}$, and consolidating ${k, l}$ into $i$ for simplicity, the 2D-3D mapping becomes:
\begin{equation}
c_{i}\begin{bmatrix}
\mathbf{u}_{i}\\
1
\end{bmatrix}=\mathcal{K}\mathbf{p}_{i}^{c}=\mathcal{K}\sum_{j=1}^{4}\alpha_{ij}\mathbf{\zeta}_{j}^{c},
\label{equ1}
\end{equation}
where $c_{i}$ is a scaling factor, $\mathbf{u}_i$ are the corner pixel coordinates, matrix $\mathcal{K}$ holds the intrinsic camera parameters (e.g. focal length and principal point coordinates), and $\mathbf{p}_{i}^{c}$ are the $3$D point coordinates with respect to the camera coordinate frame, $c$. A lemma \cite{lepetit2009ep} shows that any point in the camera space can be represented by the weighted sum of four virtual control points. These control points are denoted by $\mathbf{\zeta}_{j}^{c}\in \mathbb{R}^{3}$, $j = 0, 1, 2, 3$, and their corresponding weighting is $\alpha_{ij}$. 

We can thus reformulate \eqref{equ1} into a Homogeneous Linear Equation with respect to the virtual control points:
\begin{equation}
\mathcal{F}\mathbf{x} = \mathbf{0},
\label{equ2}
\end{equation}
where $\mathbf{x} \in \mathbb{R}^{12}$ is a stacked vector containing all the virtual control points $\mathbf{c}_{j}^{c}$. Stacked matrix $\mathcal{F}$ can be derived by \eqref{equ1} and contains all the detected points contained in $\hat{S}$.

The 2D-3D registration problem can be solved with the EPnP algorithm \cite{lepetit2009ep}. First, the algorithm determines the null space of $\mathcal{F}$, forming a vector basis through virtual control points. Second, it imposes constraints to ensure the invariance of distances between these control points across different reference frames, be it the world frame (the source points $\mathbf{p}_i$ in $\hat S$) or the camera frame. 

After the two steps of EPnP, $\hat{S}$ can be reformulated into $\Bar{S} = \{(\mathbf{p}_i^{o}, \mathbf{p}_i^{c})\}$, for $i =1,2,\cdots,k$. Here, the superscripts $o$, and $c$, denote the world frame, and camera frame, respectively. This is a conventional 3D-3D registration problem with an optimal solution:
\begin{eqnarray}
\mathcal{H} & = & (\mathbf{P}^{c})^{T}\mathbf{P}^{o}=\mathcal{U}^{T}\Lambda\mathcal{V}
\label{equ3a} \\
\mathcal{R}_{o}^{c} & = & \mathcal{U}^{T} \mathcal{W}\mathcal{V}\ 
\label{equ3b} \\
\mathbf{t}_{o}^{c} & = & \mu(\mathbf{p}_{i}^{c})-\mathcal{R}_{o}^{c}\mu(\mathbf{p}_{i}^{o})
\label{equ3c}
\end{eqnarray}
where $\mathbf{P^{c}}$ is the stacked vector of the differences between each 3D point $\mathbf{p}_{i}^{c}$ and their centroid $\mu(\mathbf{p}_{i}^{c})$, and $\mathcal{W}=$\,diag$(1.0, 1.0, |\mathcal{U}^{T}\mathcal{V}|)$.
Similarly, $\mathbf{P}^{o}$ is the stacked vector of the differences between each $\mathbf{p}_{i}^{o}$ and their centroid $\mu(\mathbf{p}_{i}^{o})$.
Matrix $\mathcal{H}$ can be decomposed via SVD, with $\mathcal{U}$, $\mathcal{V}$ leading to the optimal solution for the rotation of world frame with respect to camera frame , $\mathcal{R}_{o}^{c}$. Then, translation $\mathbf{t}_{o}^{c}$ can be found.

\subsection{Interactive force estimation}
To ensure a seamless and intuitive physical pHRI, we must consider factors like Coriolis force, friction force, and gravity force. By compensating these forces with a feedforward model, we can obtain an accurate estimation of external torque  $\boldsymbol{\tau}_\textnormal{ext}$ by:
\begin{equation}
\boldsymbol{\tau}_\textnormal{ext} = \boldsymbol{\tau}_\textnormal{raw} - \boldsymbol{\tau}_{m}.
\label{equ4}
\end{equation}
In this context, $\boldsymbol{\tau}_\textnormal{raw}$ comes from the robot's internal joint torque sensor, and $\boldsymbol{\tau}_{m}$ denotes the compensation model. With the Recursive Newton-Euler algorithm (RNEA) method \cite{KVRGIC2020103680}, the dynamics of a manipulator can be described by:
\begin{equation}
\boldsymbol{\tau}_{m} = \mathcal{M}(\mathbf{q})\mathbf{\ddot q} + \mathcal{C}(\mathbf{q, \dot q})\mathbf{\dot q} +\mathbf{g}(\mathbf{q}) + \mathcal{F_{st}}sgn(\mathbf{\dot q})+\mathcal{F_{v}}\mathbf{\dot q},
\label{equ5}
\end{equation}
In a serial robot setup, the robot's dynamic model is derived considering its inertia matrix $\mathcal{M}(\mathbf{q})$, Coriolis matrix 
$\mathcal{C}(\mathbf{q, \dot q})$, and the gravity vector $\mathbf{g}(\mathbf{q})$. Vector $\mathbf{q}$ illustrates the joint positions, $\mathbf{\dot q}$ joint velocities, and $\mathbf{\Ddot{q}}$ joint acceleration. The terms $\mathcal{F_{st}}$ and  $\mathcal{F_{v}}$ are two diagonal matrices whose elements on the diagonal correspond to the Coulomb Friction coefficient and Viscous Friction coefficient in joints, respectively. Finally, $sgn(\cdot)$ the sign function.

After obtaining the torques in joint space, we project to task space:
\begin{equation}
\mathbf{f}_\textnormal{ext}=\mathcal{J}^{T}\boldsymbol{\tau}_\textnormal{ext},
\label{equ6}
\end{equation}
where $\mathcal{J}$ denotes the robot Jacobian matrix, and $\mathbf{f}_\textnormal{ext}$ is the interactive force from both environment and operator. 

\section{Hand Eye Information Fusion}
\label{Section4}
In this section, we use a neural network acting in the Kalman filter's update part to conduct real-time hand-eye information fusion. First, the method to generate the dataset that trains the neural network is provided. Then, we discuss the network itself.

\subsection{Training dataset} 

The goal is create a dataset that is used first for hand-eye calibration and second for training the information fusion network. This dataset will not be subject to occlusions, which may, however, be present at deployment stage. Data collected through vision-based pose estimation will be referred to as $\mathbb{M}_\textnormal{cam}$, and data collected through robot forward kinematics will be referred to as $\mathbb{M}_\textnormal{kin}$. The ground-truth data distilled through filtering and registration of $\mathbb{M}_\textnormal{cam}$ and $\mathbb{M}_\textnormal{kin}$ will be referred to as $\mathbb{M}_\textnormal{gt, m}$. 


Specifically, if $t=1, 2,\cdots, T$ represent samples, $\mathbb{M}_\textnormal{kin} = \{\mathbf{z}_{b,t=1,2,\cdots,T}\ | \mathbf{z}_{b,t} = [\mathbf{p}_b^{e}\  \mathbf{q}_b^{e} ]^{T}\}$, where the end effector's position and orientation is provided with regards to the robot base, through the robot forward kinematics. Further, $\mathbb{M}_\textnormal{cam} = \{ \mathbf{z}_{t=1,2, \cdots,T} \  | \ \mathbf{z}_{t}= [\mathbf{p}_e^{o},\  \mathbf{q}_e^{o} ]^{T}\}$, where the trocar's position and orientation are provided with respect to the end effector's coordinate frame. The transformation can be calculated by the end effector's pose and the trocar's pose with respect to the camera's coordinate frame. Recall that the trocar pose is estimated through ArUco marker detection; therefore, we can use the reprojection error to filter the measurement set. 

The set with the reprojection errors is:
\begin{equation}
    \mathbb{M}_\textnormal{repr-err} = \{ \mathbf{e_{w,t=1,2,\cdots,T}}\ | \ \mathbf{z}_{t} \in \mathbb{M}_\textnormal{cam} \mapsto \hat{S} \mapsto_{\eqref{equ7}}e_{w}\}, \nonumber
\end{equation}
where,
\begin{eqnarray}
    \{e_\textnormal{proj}\} & = & \{ -\sum _{(\mathbf{u},\mathbf{p}) \in \hat{S}}||\mathbf{u} -\mathcal{K}\mathcal{R}\mathbf{p}||_{2}, \textbf{\;for each\;} \hat{S} \in \mathbb{M}_\textnormal{cam} \}, \\\label{equ7a}
    e_{w} & = & A \frac{e_\textnormal{proj}-\textnormal{min}(\{e_\textnormal{proj}\})}{\textnormal{max}(\{e_\textnormal{proj}\})-\textnormal{min}(\{e_\textnormal{proj}\})} + b,
\label{equ7}
\end{eqnarray}
where $A$ and $b$ represent a suitable gain and bias after the normalization of $e_\textnormal{proj}$. Set $\hat{S}$ is the set for pose estimation mentioned in Sec.\ \ref{Section3}A. For every $e_\textnormal{proj}$, we consider all the $\mathbf{u}$,$\mathbf{p}$ in a $\hat{S}$ in the current image frame. With this process, each measurement $\mathbf{z_t}$ in $\mathbb{M}_\textnormal{cam}$ has a re-projection error $e_w$. By using a percentage, $\epsilon$, of the measurements, i.e. those with the best quality rank, we can suppress noisy measurements and produce a filtered $\mathbb{M}_\textnormal{cam}$.

After obtaining $\mathbb{M}_\textnormal{cam}$ and $\mathbb{M}_\textnormal{kin}$, we can use the 3D-3D registration in \eqref{equ8} and \eqref{equ3b}-\eqref{equ3c} 
to get the hand-eye calibration information, namely $\mathbf{t}_{o}^{b}$ and $\mathcal{R}_{o}^{b}$. Noticeably, we replace \eqref{equ3a} with \eqref{equ8} as higher-quality measurements should carry higher weights in registration:
\begin{equation}
\mathcal{H}=\mathcal{P}_\textnormal{cam}^{T} \mathcal{W}_{r}\mathcal{P}_\textnormal{kin}=\mathcal{U}^{T}\mathbf{\Lambda}\mathcal{V},
\label{equ8}
\end{equation}
where $\mathcal{W}_{r} = diag(e_{w,t=1},e_{w,t=2}... e_{w,t=T})$ is a diagonal matrix representing the weights. Matrix $\mathcal{P}_\textnormal{cam}^{T}$ contains stacked measurements from $\mathbb{M}_\textnormal{cam}$; similarly for $\mathcal{P}_\textnormal{kin}$. From $\mathcal{U}^{T}$ and $\mathcal{V}$ we can obtain $\mathbf{t}_{o}^{b}$ and $\mathcal{R}_{o}^{b}$, which concludes hand-eye calibration.

Following this, the ground-truth measurement set becomes $\mathbb{M}_\textnormal{gt, m} = \{\mathbf{z}_{t=1,2,\cdots,T} \ |   \mathbf{z} = \textit{pose}(\mathbf{T}_{b}^{e}\mathbf{T}_{o}^{b})\}$ where $\textit{pose}(\cdot)$ returns the pose corresponds to the input transformation matrix. To use $\mathbb{M}_\textnormal{gt, m}$ for filter tuning, we incorporate the robot's motion into the dataset  
$\mathbf{\psi_t} = \begin{bmatrix}
    \delta \mathbf{p}_e^{b}, \delta \boldsymbol{\theta}_e^{b}
\end{bmatrix}^{T} $ at every time step. Denote $\delta$ as the increment operator, i.e. $\delta (\mathbf{\cdot}) = (\mathbf{\cdot})_{t+1} - (\mathbf{\cdot})_{t}$, where $t$ refers to a discrete timepoint. Therefore, the dataset for hand-eye fusion can be represented with: $M_\textnormal{gt,m}$ = $\{\mathbf{z}_{t=1,2,3...T},\mathbf{\psi_t}\}$ where $\psi_t$ comes from the robot's incremental motion which could be denoted as $\psi_t = [\delta \mathbf{p}_o^{e}, \delta \boldsymbol{\theta}_o^{e}]^{T}$ at t-th samples.

\subsection{Hand-eye information fusion network}

Trocar pose can be estimated with respect to the marker attached on the end-effector using the acquired images, detected ArUco markers, and 3D/3D correspondences. This relative transformation could be directly used for docking navigation as the error signal within a PBVS loop, since it shows how the robot should move with respect to its end-effector frame. 

The estimated pose, however, is unstable when there are inaccurate keypoint/corner detections and occlusions. Measurements from cameras and joint torque sensors must be filtered to improve robustness. Orientation, particularly when represented as a quaternion, possesses distinct non-linear properties that prevent the direct application of standard filtering techniques \cite{youn2019combined}. 

Traditional methods employ Kalman filters that are applied on the features in the 2D image space (e.g. marker corners). The 2D image filter handles the raw data from the camera, but does not explicitly considering the robot's motion (i.e. velocity) in their prediction model \cite{moghari2007point, chen2011kalman}. In addition, it is challenging to incorporate a fluctuating number of detected keypoints/corners, meaning that filters are affected by occlusions. To obtain robust pose estimation in our system, we introduce an error-state Kalman filter. 

In the following part, we name the pose $\mathbf{s}_{t}= \begin{bmatrix}
     \mathbf{p}_o ^{e},  \mathbf{q}_o^{e}
\end{bmatrix}^{T}$ ``state" and name the pose $\mathbf{z}_{t}$ ``measurement'' for simplification.
They are actually the same value but we want to distinguish whether it is an external input (measurements) or an internal variable (states) for the Kalman filter.

 We use $\delta$ to define an error state with respect to the end-effector frame, $\delta \mathbf{s} = \begin{bmatrix}
    \delta \mathbf{p}_o^{e}, \delta \boldsymbol{\theta}_{o}^{e}
\end{bmatrix}^{T}$. The representation is a $3$D vector as the angle is multiplied into the unit vector representing the axis.

To predict how the transformation from trocar to end-effector changes, we use the kinematics of the robot to represent the error state dynamics. We assume the position of the trocar barely changes within a time step. Therefore, the changes of state are only due to robot motion. As a result, the error state dynamics for prediction is:
\begin{eqnarray}
    \delta \mathbf{s}_{t} & = & \mathcal{B} \mathbf{\psi}_{t-1},
\label{equ9a} \\
\boldsymbol{\psi}_{t-1} & = & [\delta \mathbf{p}_e^{b}, \delta \boldsymbol{\theta}_e^{b}]^{T}=\mathcal{J}\mathbf{\Dot{q}} T_s,
\label{equ9b}
\end{eqnarray}
where $b$ denotes the robot base coordinate frame, and matrix $\mathcal{B} = \textnormal{diag}(\mathcal{R}_{b}^{e},\ \mathcal{R}_{b}^{e})$ is the  matrix of the system motion model that transfers the robot end-effector's motion, $\boldsymbol{\psi}_{t-1}$, to the incremental motion $[\delta \mathbf{p}_o^{e}, \delta \boldsymbol{\theta}_o^{e}]^{T}$. 

After predicting the error state, the prediction $\delta \mathbf{s_{t}}$ can be injected into the state obtained from the last update by:
\begin{equation}
\hat{\mathbf{s}}_{t|t-1} = \delta \mathbf{s_{t}} \oplus \mathbf{s}_{t-1|t-1}
\label{equ10}
\end{equation}
where $\hat{\mathbf{s}}_{t|t-1}$ is the prior value of state $\mathbf{s}$ after leveraging the prediction error state dynamics. Operator $\oplus$ is a simple add operation in the position part and an integral operation using angular velocity on the quaternion part, see \cite{sola2017quaternion}. The posterior state $\mathbf{s}_{t-1|t-1}$ is the state obtained from the last update.

The state can be fully observed via this measurement (i.e. $\mathbf{s} = \mathbf{z}$). The conventional update approach uses the covariance of measurements and models to estimate the Kalman gain. Identification of a covariance matrix is particularly challenging when the model is not fully known, and in any case requires strong insights and experimentation for fine tuning. 

Leveraging the model fitting capabilities of neural networks, one can estimate the model and identify the appropriate Kalman gains in a supervised data-driven fashion.
A Deep Kalman Filter is a convolutional neural network network that replaces the state-space-model-dependent computation of conventional Kalman filters to overcome model inaccuracies and non-linearities \cite{hovart2023deep}. We selected KalmanNet as the backbone of our developed network, adapting it to our needs. 

\begin{figure}[t]
	\centering
	\includegraphics[width=\columnwidth]{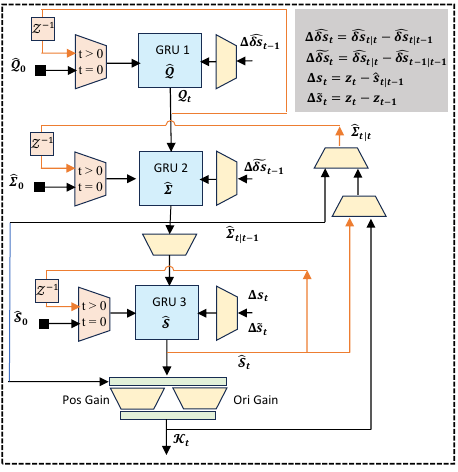}
	  \caption{The update network for KalmanNet. The inputs of KalmanNet are modified for our ESKF structure. This method infers the covariance of measurements and system models. The output of the network is the Kalman Gain containing the two independent parts (position gains and orientation gains).}
    \label{figure3}
\end{figure}

KalmanNet uses RNN to replace update calculations and carry out real-time state estimation in the same manner as model-based extended Kalman filtering \cite{revach2022kalmannet}. Compared to Transformers, RNNs are fast and appropriately precise in handling time series with $6$-$7$ dimensions. To simplify network training, KalmanNet uses Gated recurrent units (GRUs), which is a implementation of an RNN. It behaves like a long short-term memory (LSTM) module with a gating mechanism to input or forget certain features, but lacks a context vector or output gate, resulting in fewer parameters. Although our application uses the error-state Kalman Filter framework, the inputs and outputs can be reformed to be adaptive (i.e. we define the error states as inputs of the network and makes the positional part and the orientations part of outputs being independent). The amended network structure is shown in Fig.\ \ref{figure3}. 

The inputs of the network at $t = 0$ are $\mathcal{\hat{Q}}_0$, $\mathbf{\hat{\Sigma}}_0$ and $\mathcal{\hat{S}}_0$, which represent state noise covariance, predicted state moments, and updated state moments (see the moments in Kalman Filter's prediction part and update part \cite{chen2011kalman}). GRU1, GRU2, and GRU3 track three tensors ($\mathcal{\hat{Q}}$, $\mathbf{\hat{\Sigma}}$ and $\mathcal{\hat{S}}$, respectively) over time. To obtain an estimate of the quality of the first measurement from the camera, i.e. $\mathbf{z_{t=0}}$, the reprojection error of the estimated 3D corners onto the image is applied as a gain on $\mathcal{\hat{Q}}_0$. Values for $\mathbf{\hat{\Sigma}}_0$ and $\mathcal{\hat{S}}_0$ are randomly provided for initialisation. The inputs of the update network, i.e. the error states, are denoted as $\Delta \hat{\delta \mathbf{s_t}}$, $\Delta \Tilde{\delta \mathbf{s_t}}$, $\Delta \mathbf{s_t}$, and $\Delta \Tilde{\mathbf{s_t}}$ (their definitions are shown in Fig.\ \ref{figure3}). 
After the GRU blocks, the network head  ensures the output is of the structure of a gain matrix. The gain matrix is named ``Kalman Gain'' and multiplies the difference between measurements $\mathbf{z_t}$ and the predicted state $\hat{s}_{t|t-1}$ to improve state estimation (see \eqref{equ11}).

Different from the original KalmanNet, whose fully convolutional head couples all state channels, the network head in our implementation comprises two separate linear layers denoting the gain on the positional part and the orientation part, respectively. Thus, the separate layers preserve the independence of the position and orientation observations, and thereby improve training stability.

The update process suppresses the noise in estimated error states, which are now leveraged to improve the estimation of the states themselves. The update and injection process is:
\begin{equation}
\mathbf{\hat{s}_{t|t}} = \mathcal{K}_t( \mathbf{z_t - \hat{s}_{t|t-1}}) \oplus \mathbf{s_{t|t-1}}
\label{equ11}
\end{equation}
where $\mathbf{\hat{s}_{t|t}}$ is the filtered state with reduced noise in position and orientation. Then, the feedback terms including rotation matrix $\mathcal{R}_{o}^{e}$ and position $\mathbf{t}_{o}^{e}$ can be derived from $\mathbf{\hat{s}_{t|t}} = [\mathbf{p_o ^{e}}, \mathbf{q_o^{e}}]^{T}$.

\subsection{Hand-eye information fusion network loss functions}

KalmanNet is ultimately a regression that makes updated states $\mathbf{\hat{s}_{t|t}}$ approach the ground truth leveraging time series $\mathbf{z_t}, \mathbf{u_t}$. The loss function that guides training is:
\begin{equation}
C = \sum_{t=0}^{N}\gamma^{N-t} r_t,
\label{equ12}
\end{equation}
where $C$ denotes the cumulative errors of the given time sequence (short trajectories including fixed time steps truncated from $M_{gt,m}$). The error in time step $t$ is $r_t$, $N$ is the total number of steps in the given time sequence, and $\gamma\in[0, 1]$ is a discount factor that increases the impact of future rewards. The impact factor ensures the estimated trajectory will converge to ground truth.

We consider $r_t $ as the cost function of our method. It includes the reconstruction error $r_\textnormal{rc}$ and acceleration cost $r_\textnormal{acc}$ of the robotic system. The definitions of these cost functions are :
\begin{equation}
\begin{aligned}
r_t &= r_\textnormal{rc}+\beta r_\textnormal{acc}\\
&=\underbrace{\mathbf{(\hat{s}_{t|t}-s)^{T}}\mathcal{W}_{1}\mathbf{(\hat{s}_{t|t}-s)}}_{r_\textnormal{rc}}+\beta \underbrace{||\mathbf{\hat{s}_{t|t}+\hat{s}_{t-2|t-2}- 2\hat{s}_{t-1|t-1}}||}_{r_\textnormal{acc}},
\end{aligned}
\label{equ13}
\end{equation}
where $\mathcal{W}_{1}$ is a diagonal matrix that applies gain to both positions and quaternions. Due to the non-linear nature of quaternions, the algebraic subtraction described in Fig.\ \ref{figure3} is only valid for small differences. As a result, the gain assigned to the quaternion component is  lower. Finally, $\beta$ is a scalar coefficient that adjusts the influence of the acceleration term.



\section{Semi-Autonomy Controller}
\label{Section5}

In this section, we discuss the use of an optimizer for control. Our approach overcomes the requirements of traditional controllers which linearize the model at every point in the trajectory. The cost function to be optimized should be represented with respect to the feedback states (including HRI and environment-robot interaction forces). 
However, in general, it is hard to decouple the HRI and environment-robot interaction forces. 
That is, it is difficult to distinguish the forces that are applied by the human with forces that are induced from contact with the environment.
For simplification, we assume the force along the tool (z-axis of end-effector) is the HRI force, while others (including the rotational part in the wrench) are environment-robot interaction forces.

\subsection{Translational compliance criterion design}

To dock into the trocar, a controller including anisotropic axes-controlling sub-blocks is applied with respect to the axes of the end-effector frame. Equation \eqref{equ14a} represents the projection distance between the original point of the end effector and the trocar's center point with respect to the plane spanned from the x and y axes of the end effector's frame (see $e_\textnormal{xy}$ in \eqref{equ14a}). For simplicity, the superscript representing the robot base (i.e. the b in positions like $\mathbf{p}_o ^{b},  \mathbf{q}_o^{b}$) is dropped. The velocity-controlled robot motion is defined in \eqref{equ14b}:
\begin{eqnarray}
    e_\textnormal{xy} & = & \mathbf{n}_p^T\mathbf{n}_p\mathcal{R}_{e}^{T} (\mathbf{p}_{t+1}-\mathbf{p}_{o}),
    \label{equ14a} \\
    \mathbf{p}_{t+1} & = & \mathbf{v}_{t} T_s+\mathbf{p}_{e},
    \label{equ14b}
\end{eqnarray}
where $\mathbf{n}_p$ denotes a projection matrix that selects the positionally controlled subspace from the whole task space. In our case, to ignore position control along the z-axis, we set the projection matrix to $[1.0, 1.0, 0.0]^{T}$. Matrix $\mathcal{R}_{e}$ denotes the rotation from the robot base to the end effector, and $\mathbf{p}_{t+1}$ indicates the future position command. 

The position command can be obtained from feedback from the robot's current position of the end effector $\mathbf{p}_{e}$ and a desired velocity command $\mathbf{v}_{t}$ at time t, with \eqref{equ14b} showing stepwise integration. The desired velocity $\mathbf{v}_{t}$ can be calculated using the optimizer mentioned in Section IV.D. To align the trocar, the projection distance $e_\textnormal{xy}$ should be controlled to zero and this is the goal of velocity generated by the optimizer. 



To compliantly align the end-effector with the trocar, we consider the compliance during the docking as a large interactive force that should be avoided for safety, which leads to the new aligning error considering external force:
\begin{equation}
\begin{aligned}
    e_\textnormal{xy,f} &= e_\textnormal{xy,ext} + e_\textnormal{xy}\\
    &=\underbrace{\mathbf{n}_p^T\mathbf{n}_p\mathbf{B}_\textnormal{x,y}[\mathbf{I}_{3\times3}\;\mathbf{0}_{3\times3}]\mathbf{f}_\textnormal{ext}}_\textbf{force compliance error}+ \underbrace{\mathbf{n}_p^T\mathbf{n}_p\mathcal{R}_{c}^{T}(\mathbf{p}_{t+1}- \mathbf{p}_{o})}_\textbf{projection error from \eqref{equ14a}},
    \label{equ15}
\end{aligned}
\end{equation}
where the error $e_\textnormal{xy,f}$ comes from the force compliance error $e_\textnormal{xy,ext}$ and the projection error $e_\textnormal{xy}$. To formulate the force compliance error, admittance control is used. Therein $\mathbf{B}_\textnormal{x,y}=diag(\begin{bmatrix}
    B_{x}, B_{y}, 0
\end{bmatrix})$ denotes a diagonal matrix including x and y part admittance: $B_{x}$ and $B_{y}$, and $\mathbf{f}_\textnormal{ext} \in \mathbb{R}^{6}$ is the measured force which is acted on end-effector during the task.

\subsection{Rotational compliance criterion design}

The main task concerning rotation is to align the tool's axis to the trocar's axis. To this end, we build an error on the rotation part considering the inner product of the z-axes of the end-effector and trocar coordinate frames:
\begin{equation}
    e_{R,f} = \mathbf{\xi}_{o,z}^{T}\mathbf{\xi}_{t+1,z}-1.0,
    \label{equ16}
\end{equation}
where $\boldsymbol{\xi}_{o,z}$ is the third column of rotation matrix $\mathcal{R}_{o}$. The optimized variable here is $\boldsymbol{\xi}_{t+1,z}$, which is the third column of the predicted orientation $\mathcal{R}_{t+1}$. We also have a kinematic constraint: $\mathcal{R}_{t+1} = (I+T_s \Omega_{t})\mathcal{R}_{o}$, wherein  
$\boldsymbol{\Omega}_{t}$ denotes the desired angular velocity of the end effector.
We introduce $\mathbf{f}_{ext}$ to conduct compliance control in rotation with the criterion that follows:
\begin{equation}
\begin{aligned}
    \boldsymbol{\xi}_{t+1,z} &= \boldsymbol{\xi}_{t,z} + (e_{r} + e_\textnormal{r,ext})\mathbf{c_z}\\
    &= (\underbrace{T_s \boldsymbol\Omega_{t}\mathcal{R}_{o}}_\textbf{rotation error}  + \underbrace{T_{s}\boldsymbol\Omega_{ext}\mathcal{R}_{o}}_\textbf{torque compl.~err.} + \mathcal{R}_{o})\underbrace{[0\ \ 0\ \ 1]^{T}}_\textbf{z axis},
    \label{equ17}
\end{aligned}
\end{equation}
where $\boldsymbol{\xi}_{t+1,z}$ is the desired z axis at time step $t$. The z axis can be represented by the last z axis updated by adding the errors, including rotation error, from t to t+1, $e_{r}$, and the error regarding compliance, $e_\textnormal{r,ext}$. The errors are projected to the z axis by $\mathbf{c_z}$. In the rotation error part, $\boldsymbol\Omega_{cmd}$ is a skew-symmetric $3\times 3$ matrix derived from the angular velocity vector of the end effector. In the torque compliance error part, we use $\boldsymbol\Omega_\textnormal{ext}=(\mathcal{B}_{R}[\mathbf{0}_{3\times3}\;\mathcal{I}_{3\times3}]\mathbf{f}_\textnormal{ext})^{\wedge}$ to serve as a virtual angle velocity matrix. Matrix $\mathcal{B}_{R}$ is the compliance matrix, defined by $\textnormal{diag}(B_{rx},B_{ry},0)$. In $\mathcal{B}_{R}$, the z channel (the third element in the diagonal) is zero because the rotation along z is not relevant to the docking task. The angular velocity is added to the velocity command $\boldsymbol\Omega_\textnormal{cmd}$. 

\subsection{Co-manipulation control}
To make guidance intuitive, an admittance controller is applied along the direction of the tool's axis. The guidance force error minimised in the optimizer is:
\begin{equation}
    e_{z,f} = \underbrace{\mathbf{n}_f^{T}\mathbf{n}_f }_\textbf{projection to force subspace}(\underbrace{\mathcal{B}_z[\mathbf{I_{3\times3}}\;\mathbf{0}]\mathbf{f}_\textnormal{ext}}_\textbf{target vel by admittance}-\mathbf{v}_t),
    \label{equ18}
\end{equation}
where $e_{f}$ denotes the force error during HRI, and $\mathbf{n}_f = \mathbf{\xi_{e,z}}$ denotes the force subspace, which is the z vector in the rotation matrix of the end effector frame with respect to the base frame. Symbol $\mathcal{B}_z = \textnormal{diag}(0, 0, B_z)$ describes the admittance value for the force subspace, while  $\mathbf{v}_t$ here denotes the desired velocity and is also the command to the co-manipulated serial robot.

\subsection{Optimizer implementation}

We weight all cost functions to reflect their relative importance:
\begin{equation}
r_{q}=\underbrace{w_{z,f}||e_{z,f}||_{2}^{2}}_\textbf{translational compliance}+\underbrace{w_\textnormal{xy,f}||e_\textnormal{xy,f}||_{2}^{2}}_\textbf{rotational compliance}+\underbrace{w_{R,f}||e_{R,f}||_{2}^{2}}_\textbf{co-manipulation}
    \label{equ19}
\end{equation}
where $w_{z,f}$,$w_\textnormal{xy,f}$, and $w_{R,f}$ represent the corresponding weights. 

Although we could directly optimize the model using Nlpsol\cite{biegler2010nonlinear}, this approach would lack a safety guarantee and the optimizer could fail to find a solution or cause the robot to run into a singular configuration. We address this by introducing velocity limits both in Cartesian space and joint space, with the optimization becoming:
\begin{subequations}\label{eq20}
\begin{align}
\mathop{min}_{\mathbf{\dot{q}_t}}\;& r_q (\mathbf{q}_e,\mathbf{\hat{s}},\mathbf{v}_{t},\mathbf{\Omega}_{t}) + w_j ||\mathbf{\dot q}_t||_{2}^{2},  \label{equ20a} \\
\text{s.t.} \;&\mathbf{v}_{t}:=\mathcal{J}_{l}(\mathbf{\dot{q}})\mathbf{\dot q}_t, \label{equ20b}\\ 
&\mathbf{\Omega}_{t}:=(\mathcal{J}_{\Omega}(\mathbf{\dot{q}})\mathbf{\dot q}_t)^{\wedge}, \label{equ20c}\\ 
&\mathbf{v}_{t}<\mathbf{\Bar{v}}_{t,u}
\label{equ20d}\\
&\mathcal{J}_{\Omega}(\mathbf{\dot{q}})\mathbf{\dot q}_t<\mathbf{\Bar{\omega}}_{t,u},
\label{equ20e}
\end{align}
\end{subequations}
where $w_j$ weights the velocity-minimum criteria term to avoid singularities. Terms $\mathcal{J}_{l}(\mathbf{\dot{q}})$ and $\mathcal{J}_{\Omega}(\mathbf{\dot{q}})$ are the translational and rotational parts of Jacobian matrix. 
The superscript $\wedge$ is an operation on SO(3), which transfers a 3-channel vector to its skew-symmetric matrix, and vice versa for $\vee$. 
The upper bounds for linear velocity $\boldsymbol{\Bar{\omega}}_{t,u}$ and angle velocity $\boldsymbol{\Bar{\omega}}_{t,u}$ are given through trials, and are hyperparameters.

The optimization problem defined in~\eqref{eq20} is implemented using OpTaS, a library for optimization-based planning and control~\cite{mower2023optas}. Specifically, we use the \textit{sqpmethod} as the solver.
As the \textit{sqpmethod} 
proceeds by successively solving linearized sub-problems,
it performs well when the the initial point $\mathbf{q}_e$ is far from the singularity point. 

The previous converged solution serves as the starting point of the optimization. To seed the first optimization iteration, we provide the optimizer with a  solution only considering the part task of guiding which is related to the HRI guiding force along the z-axis, i.e., \eqref{equ18}. 

\begin{figure}[t]
	\centering
	\includegraphics[width=\columnwidth]{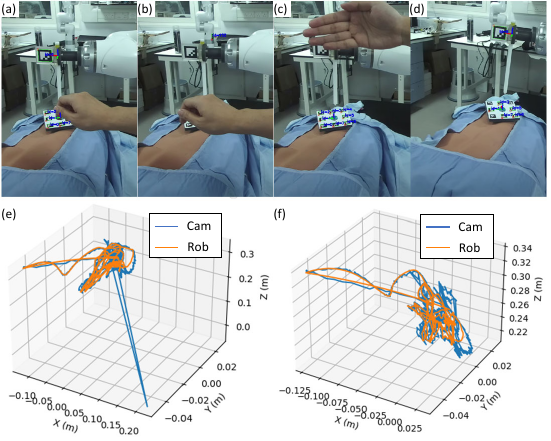}
	  \caption{Illustration of disturbances in the dataset and results of the hand-eye-calibration-based robot's motion (Rob lines) which serves as ground truth (a) Part Occlusion of the marker of the trocar; (b) Full Occlusion of the marker of the trocar; (c) Occlusion of the marker of the end effector; (d) Camera real-time movement; (e) Example 3D plot of measured positions (Cam) and Rob explaining robustness of ground truth in the presence of a single outlier; (f) Example 3D plot of measured positions (Cam) and Rob explaining robustness of ground truth while frequent outliers exists.}
    \label{figure4}
\end{figure}

\section{Data Collection and Self-Supervised Task Construction}
\label{Section6}

As we use supervised learning to identify the parameters of the filter that supports marker-based pose estimation, emphasis is placed on collecting the appropriate dataset. As usual, the collected dataset will be be separated in the non-overlapping split of training set (60 \% of data), validation set (20 \% of data), and test set (20 \% of data). Collected trajectories/data have the format $\mathbb{M}_\textnormal{gt,m}$ = \{$(\textbf{z}_{t}, \textbf{u}_{t})_{t=1,2,3..., T}$$| \textbf{z}_{t}= \begin{bmatrix}
     \textbf{p}_{o,t}^{e},  \textbf{q}_{o,t}^{e}
\end{bmatrix}, \textbf{u}_{t} = \begin{bmatrix}
     \delta \textbf{p}_{e,t} ^{e},  \delta \boldsymbol{\theta}_{e,t}^{e}
\end{bmatrix}$\}. 
The discrete sampling timepoint is denoted by $t$, ranging from 1 up to the maximum time, T. The subscript, such as the $t$ in $p_{o,t}^{e}$, indicates the specific timepoint. For ease of explanation, we will omit the $t$ notation in subsequent descriptions. The dataset is automatically labeled by a dedicated hand-eye calibration method (see Sec.\ \ref{Section4}A).

\subsection{Data collection}

All trajectories/data were gathered by through human-robot interaction shared control. The human operator was tasked to align the tool with the trocar for docking. Intentional partial and full blocking of the ArUco markers took place to evaluate the robustness of the framework, see Fig.\ \ref{figure4}(a-c). The camera was also manually relocated while still allowing for visualisation of the workspace, see Fig.\ \ref{figure4}(d). The trocar, however, was stable throughout the operation to ensure consistency in the ground truth. 

For each trajectory, we recorded the robot's end effector motion, $\mathbf{x}_r = [\textbf{p}_b^{e}, \textbf{q}_b^{e}]$, and the transformation pose of the trocar with respect to the end-effector $\mathbf{x}_c = [\textbf{p}_o^{e}, \textbf{q}_o^{e}]$.  In this context, the subscript $e$ denotes the frame of the robot's end effector, $b$ indicates the frame associated with the robot's base, $o$ specifies the frame of the trocar, and $c$ represents the frame of the camera. 

\subsection{Construction of self-supervised tasks}

First, hand-eye calibration must be carried out to identify the homogeneous transformation between the camera frame and the robot end effector frame, with two key features: 
\begin{itemize}
    \item The calibration registers the transformation from the trocar frame to the robot's base frame as their related position and orientation are fixed during the demonstration.
    \item Using \eqref{equ3a}-\eqref{equ3c} allows the estimation of reprojection errors. Using these errors, we select the top 20\% good quality measurements ($\epsilon = 20\%$) as input to the registration. The gain and bias of $e_{w}$ in \eqref{equ7} were set to $A = 0.8$ and $b = 0.2$.  
\end{itemize}


After transferring $\mathbf{x}_e$ to the trocar frame, we can get the ground truth trajectory of the robot with respect to the camera frame. An example trajectory from our recorded data is shown in Fig.\ \ref{figure4}(e-f), which also illustrates that the maximum distances between the ground truth and observations are related to outliers. The presence of outlines shows that the process may fail when occlusions/camera movement occurs. The maximum outlier distance in all $11$ trajectories is $0.5994\,$m, with the average/variance being $0.44 \pm 0.12 $\,m. The average error between the proposed ground truth and the observations is $0.012 \pm 0.005$\,m, but outliers skew this value. To give the estimation of actual steady error, we used the Mean Shift to find the error, which is the point in 3D Cartesian space with the maximum local density (robot being steady there). Bandwidth $0.1$ is set up for Mean Shift and the results of steady error is $0.39 \pm 1.95\times10^{-2}$\,mm, which is reasonable considering the dimensions of the trocar, workspace etc.

\section{Experiments and Discussion}
\label{Section7}

This section presents a series of experiments designed to evaluate the KalmanNet-based state estimation and docking in a phantom.

Initially, we assess the system's spatial awareness and accuracy in pose estimation. Subsequently, we examine its precision in a controlled docking task, followed by testing its robustness in a complex, real-world-like phantom environment. These experiments collectively aim to demonstrate the system's effectiveness in navigating and performing the task of semi-autonomous docking.

\subsection{Training and evaluation of information fusion network}

The KalmanNet-based hand-eye information fusion network is a recursive architecture with both external recurrence and internal GRUs. This architecture is suitable for back-propagation through time (BPTT) \cite{werbos1990backpropagation}. As the original KalmanNet work suggested, the architecture can be unfolded across time with shared network parameters, so that a forward and backward gradient estimation pass through the network. To make the model stable, we consider the truncated BPTT algorithm. There, long trajectories (e.g. N = $7000$ time steps) are each divided into multiple fixed and short trajectories (e.g. N = $150$ time steps). 
For training we use the Adam optimizer with a learning rate of 0.001.
L2 normalization is used to mitigate overfitting by penalizing large weights and encouraging the model to develop a generalized solution.

In our study, we assume that we know the system dynamics, e.g. from \cite{tian2024excitation} or the manufacturer. As a result, we implemented a conventional Extended Kalman Filter (EKF) as the baseline of our state estimation framework. Furthermore, we also utilized the original formulation of KalmanNet, to have a data-driven benchmark against which to compare our proposed method. A comprehensive series of experiments was conducted on the validation set to assess the performance and robustness of our implemented filters in comparison to the baselines. The results are shown in Table \ref{table1}.


\begin{table}[h]
\centering
\begin{tabular}{l c|c|c|c|c}
\toprule
 & \multicolumn{5}{c}{ Performance test } \\
\cmidrule{2-6}
Method& value& $e_{pos}$ & $e_{ori}$ & $a_{pos}$ & $a_{ori}$  \\
\midrule
Observation & \( \hat{\mu} \) & 0.486 & 2.046 & 0.287 & 1.399  \\
          & \( \hat{\sigma} \) & \(\pm\)0.999 & \(\pm\)2.847 & \(\pm\)0.475 & \(\pm\)1.294  \\
\midrule
ESKF & \( \hat{\mu} \) & 0.298 &  1.821& 0.051 & 1.150  \\
    & \( \hat{\sigma} \) & \(\pm\)1.661 & \(\pm\)2.365 & \(\pm\)0.015 & \(\pm\)0.004 \\
\midrule
EKF  & \( \hat{\mu} \) & nan & nan & nan & nan \\
    & \( \hat{\sigma} \) & \(\pm\)nan & \(\pm\)nan & \(\pm\)nan & \(\pm\)nan  \\
\midrule
KF  & \( \hat{\mu} \) & nan & nan & nan & nan  \\
    & \( \hat{\sigma} \) & \(\pm\)nan & \(\pm\)nan & \(\pm\)nan & \(\pm\)nan  \\
\midrule
KalmanNet  & \( \hat{\mu} \) & nan & nan & nan & nan \\
(Original)    & \( \hat{\sigma} \) & \(\pm\)nan & \(\pm\)nan & \(\pm\)nan & \(\pm\)nan  \\
\midrule
ES KalmanNet & \( \hat{\mu} \) & 0.417 & 0.443 & 0.053 & 0.550  \\
(Coupled gain)    & \( \hat{\sigma} \) & \(\pm\)0.168 & \(\pm\)0.675 & \(\pm\)0.014 & \(\pm\)0.78  \\
\midrule
Hand-eye calib.  & \( \hat{\mu} \) & 0.311 & 4.833 & \textbf{0.050} & 0.303  \\
    & \( \hat{\sigma} \) & \(\pm\)1.871 & \(\pm\)12.341 & \(\pm\)\textbf{0.014} & \(\pm\)0.000  \\
\midrule
ES KalmanNet & \( \hat{\mu} \) & \textbf{0.075} & \textbf{0.408} & 0.053 & \textbf{0.500}  \\
(our method)          & \( \hat{\sigma} \) & \(\pm\)\textbf{0.191} & \(\pm\)\textbf{0.670} & \(\pm\)0.016 & \(\pm\)\textbf{0.001}  \\
\bottomrule
\end{tabular}
\caption{Performance in both reconstruction and smoothing sections: $\hat{\mu}$ infers the mean value of results and $\hat{\sigma}$ refers to the standard derivation.}
\label{table1}
\end{table}


In Table\ \ref{table1}, $e_\textnormal{pos}$ denotes the position reconstruction error, calculated via the L2 norm. The error $e_\textnormal{ori}$ is the quaternion reconstruction error. The two errors are the positional part and orientation part of $r_\textnormal{rc}$ in \eqref{equ13}. Similarly, $a_\textnormal{pos}$ and $a_\textnormal{ori}$ are the positional part and orientation part of $r_\textnormal{acc}$ in \eqref{equ13}.

\textbf{Observation} are the real-time unfiltered measurements of the robot's trajectory with respect to the camera frame as calculated by the AruCo markers. \textbf{ESKF}, \textbf{EKF}, and \textbf{KF} refer to the error-state Kalman filter, extended Kalman filter, and Kalman filter, respectively. 
\textbf{KalmanNet (Original)} relates to the KalmanNet mentioned in \cite{revach2022kalmannet}. We directly take positions and quaternions as input. \textbf{ES KalmanNet (Coupled gain)} leverages the KalmanNet structure mentioned in \cite{revach2022kalmannet} but we take the error state of positions and quaternions as input.
\textbf{Hand Eye calibration} refers to using 1-2 prior measurements to do calibration to build the tranformation from camera to the robot base, and then inferring the future state via the robot's kinematics. 
 \textbf{ES KalmanNet (our method)} is the updated KalmanNet structure shown in Fig.\ \ref{figure3}. It corresponds to the modified KalmanNet, wherein position and orientation gains are decomposed in the update stage. 

For the shake for simplification, we will use the name ``our method" to refer to \textbf{ES KalmanNet} (our method), and use ``KalmanNet'' to refer to \textbf{ES KalmanNet} (coupled gain).

The results in Table\ \ref{table1} show that our method has the least error in $e_\textnormal{pos}$ and $e_\textnormal{ori}$. It also has the best smoothing performance in $a_\textnormal{ori}$. Although it is the 3rd best model regarding $e_\textnormal{pos}$, there is only a small differences to the best performing one (mean: 0.05 v.s. 0.053 (ours); std: 0.014 v.s. 0.016 (ours)).
Noticeably, the nan mentioned in the performance test shows the error exceeded the float format in python, signifying the worst of performances. 

\begin{figure}[t]
	\centering
	\includegraphics[width=0.95\columnwidth]{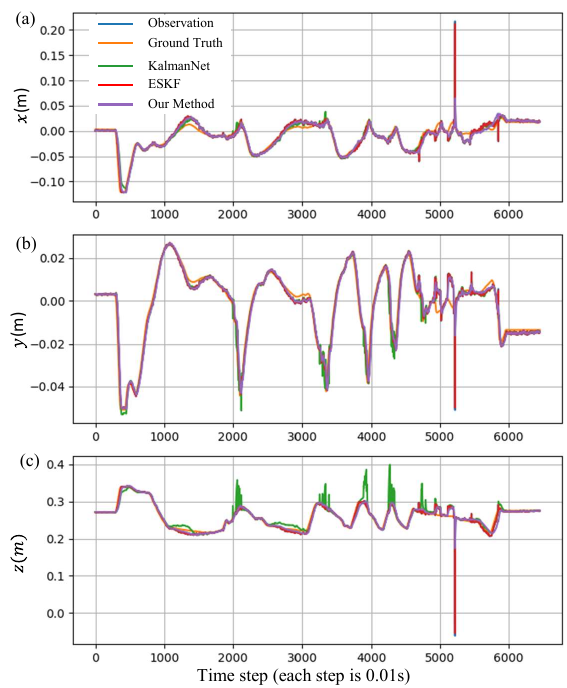}
	  \caption{Long-time series fusion experiment (position component). The plots show the predicted positions along the x, y, z axes.}

    \label{figure5}
\end{figure}

\begin{figure}[t]
	\centering
	\includegraphics[width=\columnwidth]{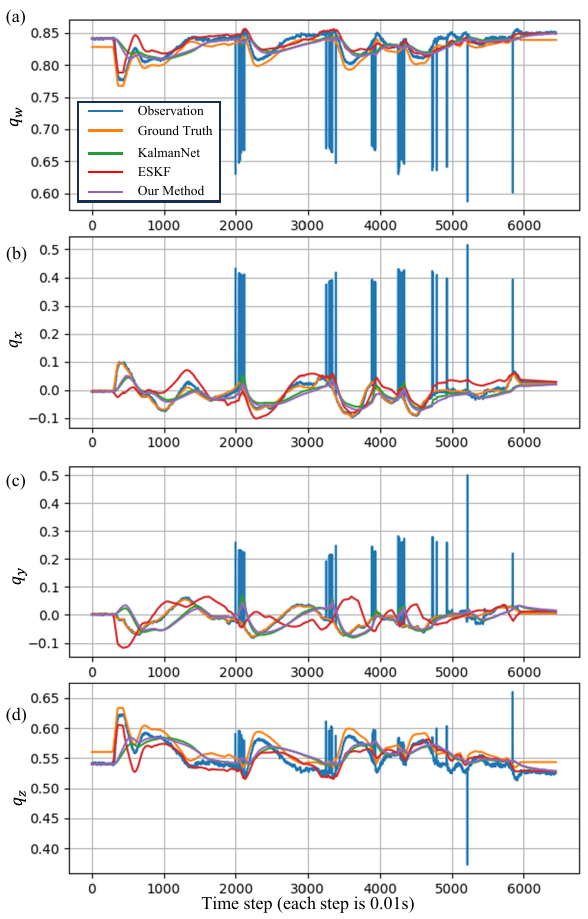}
	  \caption{Long-time series fusion experiment (orientation component). The qw, qx, qy, and qz channels of the predicted quaternion representations are provided.}
    \label{figure6}
\end{figure}

Recall that the models were trained on truncated, short, and fixed-step trajectories but it is crucial to evaluate the method's scalability and its ability to handle extended sequences. Therefore, in the test dataset, the proposed method is applied to optimize long-time observations ($T \geq 6000$ time steps, around 200$\;s$ according to 30Hz frequency of ZED camera), thereby assessing the model's ability to maintain accurate predictions over a prolonged duration. The performance of the assessed methods, shown in Fig.\ \ref{figure5} (position) and Fig.\ \ref{figure6} (orientation), support the conclusions arising from observing Table\ \ref{table1}. 

The average value of loss function $C$ described in \eqref{equ12} was 0.00038 compared to the loss of observation 0.00049. The value achieved by ESKF was 0.000693 and the value achieved by the original KalmanNet was 0.00044.
We also averaged the $r_c$ (position errors $e_\textnormal{pos}$ and orientation errors $e_\textnormal{ori}$ in \eqref{equ13}) of every sampled point in a trajectory (around 6500 steps) to compare different methods. As the methods that do not use error state tricks get nan results, we drop them off in the visualized results. 
The results in Fig.\ \ref{figure5} and Fig.\ \ref{figure6} show that our method has the least error in orientation $e_\textnormal{ori} = 0.036 \pm 0.018$. It also reached the result in positions ($e_\textnormal{pos} = 0.0083\pm 0.0079$), which is only second to the result of ESKF ($e_\textnormal{pos} = 0.0067\pm 0.0068$). Both of the methods are good enough in position error, and there is not much differences between them.
However, ESKF failed in quaternions and only obtained $e_\textnormal{ori} = 0.059\pm0.036$, which had the largest error among the methods. The visualized results were also shown in Fig.\ \ref{figure6} where the red line is far away from the ground truth. 
The KalmanNet method had positional errors $e_\textnormal{pos} = 0.0096 \pm 0.012$ and $e_\textnormal{pos} = 0.041\pm 0.022$. Both performances were good enough; however, the position and orientation performance improve after decoupling them.

Our results demonstrate that transformed robot trajectories given by hand-eye calibration could serve as the ground truth of KalmanNet which requires smooth, accurate pose states as the ground truth. Compared to the direct camera observations, it had nearly no outliers as the robot's sensor can always accurately observe the robot's motion and does not have issues with partial occlusions. We aligned the robot's trajectory and the camera's observation. Then we obtained the positional difference between them. The statistics study showed large outliers (maximum 0.5994 m and average/standard derivation: $0.44 \pm 0.12m$) existed in the camera observations. Although outliers exist, the ground truth is robust and stable. The steady error ($0.39 \pm 1.95 \times 10^{-2} mm$) was small enough for our application. This suggests that the generated ground truth can be used to train a real-time hand-eye information fusion model (KalmanNet-based filter in our cases).

We finally evaluated latency and processing throughput under various operational conditions to ensure that the algorithm's execution time aligns with the timing constraints inherent in robotic surgery applications. Tests were run on a PC equipped with AMD 5900HS, 16GB, and an NVIDIA GTX 1650 (RAM 4GB). The average inference time of our method was $2.60\,$ms with a standard derivation of $0.128\,$ms. The CPU version of the network led to slightly increased execution time of $5.79\,$ms with the standard derivation being $0.119\,$ms. In either scenario, the computational cost was lower than the camera sampling time ($33\,$ms), and had a minimal overhead in the execution of the control loop needed for HRI.

Compared to the findings of offline hand-eye calibration, other filters, and observation-only methods, our method obtained the best results on the reconstruction error of position $e_\textnormal{pos}$, the reconstruction error of orientation $e_\textnormal{ori}$, and the representation of orientation acceleration $a_\textnormal{ori}$. In the representation of positional acceleration $a_\textnormal{pos}$, our method obtained the third-best result and is almost the same as the first place in terms of mean and variance. The time-costing performance suggested the method could be integrated into the robotic system as part of a real-time controller and won't cause delays. Contrary to the findings of the original filter, our results showed quaternion following could be a challenge for the GRU network/original Kalman filter as the quaternion was non-linear, with inherent constraints, and not an injective representation on the SO3 group. Therefore, tracking the difference (error state) was practical in our scenario. Our experiments showed that all the error-state-free filters failed to handle orientation tracking and gave nan in the performance test. With the BPTT algorithm, our findings in the long-term test suggested that although the time sequence was fixed and short, the GRU could learn the right features to handle longer time series. This reinforces the KalmanNet could be extended to the longer time-series pose tracking than the fixed time sequences we used in the dataset. The original KalmanNet cannot handle quaternions (will be unstable during training); however, the ESKalmanNet (coupled output) performs better. Issues in handling quaternions can  cause the jitter in the positional plot (in Fig.\ \ref{figure5}). Although the decoupled KalmanNet has a similar performance to coupled KalmanNet in orientation performance, it tracks better the position. This shows that error state separation is also beneficial for deep filters.
However, we can infer that the proposed fusion-based methods are dependent on the initial guess of the target position. In our case, the initial phase does not have significant occlusions or motion blur, which explain its good performance. 



\subsection{Docking experiment on phantom}

In this part, we leveraged our model to conduct docking tasks. As we mentioned in Sec.\ \ref{Section2}, surgeons could push or pull the robot with a little force along the tool's axis. The force along other directions was treated as an environment-robot interaction force and the robot adopted a compliant behavior reacting to these forces. 

The parameters mentioned in \eqref{equ15},\eqref{equ17}, and \eqref{equ18} were derived based on preliminary experiments: $B_x = 0.02$, $B_y = 0.02$,$B_z = 0.1$, $B_\textnormal{rx} = 0.2$, $B_\textnormal{ry} = 0.2$. The sampling time was set according to the communication frequency between the PC and control box of the KUKA robot, i.e. $T_s = 0.01\,$s ($100\,$Hz). The weights in \eqref{equ19} were $w_{z,f} = 0.5$, $w_{xy,f} = 0.1$, and $w_{R,f} = 0.1$. The $w_j$ mentioned in \ref{equ20a} was set to $0.1$. The velocity limits in translation velocity and angle velocity in \eqref{equ20d} and \eqref{equ20e} were defined as $\mathbf{\Bar{v}_{t,u}}=\begin{bmatrix}
    0.03, 0.03, 0.03
\end{bmatrix}^{T}$, $\mathbf{\Bar{\omega}_{t,u}}=\begin{bmatrix}
    0.3, 0.3, 0.3
\end{bmatrix}^{T}$.

Co-manipulation was evaluated through 10 experiments, evenly split between a control group and a test group. Both groups experienced the same shared controller. The test group had ``access'' to the KalmanNet-based estimating optimization. On the other hand, the control group experienced a classically designed error-state Kalman filter, which was fine-tuned to the best of our ability. We considered different situations in the error-state Kalman filter, like only using observations, and only using model predictions (mentioned in the Introduction part) in the fine-tuned process. 

In terms of control-group selection, the docking trials for verification were carried out 10 times to reach high docking success rates. We ultimately selected a combination of observation and prediction methods rather than only using observations or predictions, as this approach demonstrated superior performance, achieving a 100\% success rate in docking trials compared to a 90\% successful rate for only observation and a 60\% successful rate for only prediction.

The interaction force and position disturbance served as the criteria to evaluate the performance of the docking procedure. The interactive force and position disturbance refer to the horizontal position error and the horizontal force when the tool was inserted into the trocar. These criteria ensured that any variables potentially leading to incision tearing can be attributed to the observed interactions and positional jitters.

\begin{figure}[t]
	\includegraphics[width=\columnwidth]{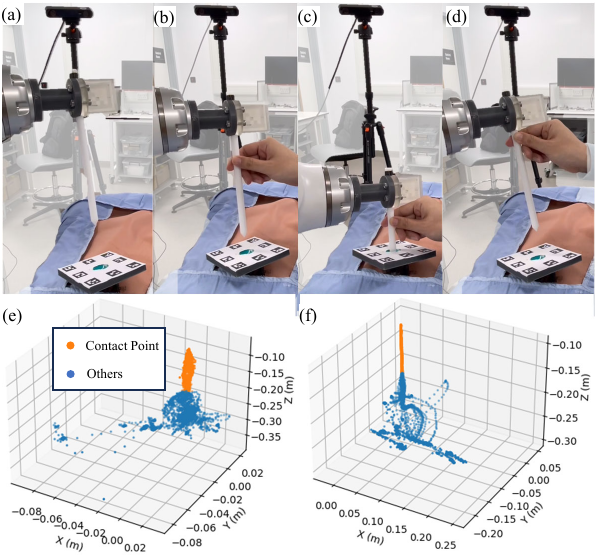}
	  \caption{Experimental setup and data visualization for trocar-robot interaction study. (a-d) Sequential stages of the trocar insertion procedure conducted by the robotic arm. (e) 3D scatter plot showing the dispersion of position measured at all points with emphasis on the contact points (orange) and other points (blue) in the control group. (f) The same type of data as in (e) but in the test group.}
    \label{figure7}
\end{figure}

The co-manipulation process along with the robot's corresponding trajectories was depicted in Fig.\  \ref{figure7}, with subplots (a-d) showing the sequential stages of interaction and subplots (e-f) illustrating the trajectories. Notably, the depicted positions indicate the trocar's location relative to the end-effector frame. The orange points are the points below the ArUco marker's surface. During interaction, the docking performance is worse if the degree of dispersion of the orange point cluster is larger. Qualitatively, the width of orange point cluster in Fig.\ \ref{figure7}(e) is more disperable than in Fig.\ \ref{figure7}(f).

In order to quantify interaction and jitters, we employed a PCA-based algorithm to assess the level of position/force dispersion in the horizontal plane. Specifically, we selected trajectory points with z-values above a predefined threshold of $-0.20\,$m, indicative of trocar-robot interaction during contact. The ideal trajectory should strictly adhere to Pfaffian constraints, mirroring the straight channel of the trocar. Ideally, all trajectories would resemble a straight line, with deviations perpendicular to this line indicating unfavorable disturbances. This line was well defined in terms of positional dispersion, represented as $[0\;0\;1]$. However, regarding force dispersion including HRI force which was vague, different trials might deviate from this ideal line. By applying PCA to these forces, we determined the primary direction of exerted force, with the principal component representing the line's direction. Subsequently, every point could be projected onto this line, and the distance between the point/force and the line was defined as position/force dispersion, respectively. 

The position and force dispersion can qualitatively evaluate the shear stress on patients' tissue during operation, which may lead the skin open around the trocar shaft \cite{blinman2010incisions}. The control group's mean position dispersion was $2.47\pm1.22$\,mm, while the test group achieved $1.23\pm0.81$\,mm, i.e. reduced by almost 50\%. As far as force dispersion is concerned, the control group's value was $1.15\pm0.97$\,N, compared to the test group's $0.78\pm0.57$\,N, again, an almost 50\% reduction. Finally, the maximum force dispersion of the control group was $5.41\,$N while the value of the test group was $2.37\,$N.

The docking experiments showed that the implementation of the controller, pose estimation, and hand-eye information fusion network was stable in real-time and real-world systems. In the error-state Kalman Filter part, we chose the ESKF including both robot kinematics and observation as our baseline as the docking successful rate using the fusion is the highest. Although the self-supervised dataset was derived from a static trocar according to the requirements of trocar registration, the proposed KalmanNet filter performed well on a real docking scene where the trocar could move according to manipulation or external disturbance. 

All the docking experiments (including the control group and test group) were successful. Considering the potential damage that co-manipulation could bring, we introduced force and positional dispersion testing experiments to indicate the safety factors that may be considered in potential applications.  Compared to the control group which leveraged the error-state Kalman Filter, the proposed method had less position and force dispersion. The results among mean value, standard derivations, and maximum dispersion showcase the superiority of the proposed approach to the baseline.

\section{Conclusion and Future Work}
\label{Section9}

In this work, a semi-autonomy system was developed for the initial docking phase of a laparoscopic operation.
An RGB camera was used to obtain measurements of the pose of the trocar through which the laparoscope should be inserted. An occlusion-robust pose estimation and hand-eye information fusion approach was presented to tackle noise and outliers. Our approach was trained on a self-supervised dataset, and demonstrated superior performance over a variety of state-of-the-art and conventional state estimation approaches.

The real-world phantom experiment showed that the proposed method achieved position dispersion of $1.23 \pm 0.81$\,mm compared to $2.47 \pm 1.22$\,mm for the control group and force dispersion $0.78 \pm 0.57$\,N versus $1.15 \pm 0.97$\,N for the control group, which is expected to bring less shear stress on incisions of patients\cite{blinman2010incisions}.

We highlight two main limitations of our proposed approach. 
First, our method has the dependency on markers, which may be contaminated during the procedure.
Whilst for certain procedures the risk of contamination is high, the risk in keyhole surgery is negligible since the main operation is contained within the patient's abdomen. Also, the occlusions on the markers of robot's end effector is not considered, which may gradually affect the success rate of docking as the number of medical assistants increases.
Second,  we distinguished HRI and environment-robot interaction forces only by their direction. This dosen't make sense when the trocar cannot be described by a quasi-static process (e.g. breathing, heart beating). 
In our experiments, this did not pose a challenge since we were able to position the camera in a convenient location.
This may not be the case, however, in general surgical operating theatre settings.
Re-calibration of the hand-eye transform is possible however could be tedious or lead to an undesirable duration of the procedure.
An alternative solution is to attach the camera to an external arm and employ dynamic camera repositioning (e.g.~\cite{Gleicher-RSS-19}).

In the future, we will integrate the marker-less pose estimation\cite{wen2023foundationpose} of both the trocar and the robot into the fusion model. The fusion model could incorporate the prediction of HRI force and environment-robot interaction into the ``prediction'' part of the KalmanNet. This may help to distinguish the task allocation and obtain more accurate trocar's pose as it leverages the force information (dynamics) of environment. 
Furthermore, whilst work addresses the problem of laparoscopic instrument docking, it has the potential to extend to wider healthcare applications like percutaneous puncture \cite{shi2022synergistic}, and nasopharyngeal swab \cite{wang2020design}.



\bibliographystyle{IEEEtran}
\bibliography{references}

\end{document}